\documentclass[lettersize,journal]{IEEEtran}
\usepackage{algpseudocode}
\usepackage{graphics}
\pdfoutput=1
\usepackage{enumitem} 
\usepackage{float}
\usepackage{textcomp}
\usepackage{amsfonts,amssymb}
\usepackage{bbm} 
\usepackage{algorithmicx,algorithm}
\usepackage{booktabs}
\usepackage{setspace}
\usepackage{amsmath,boxedminipage}
\usepackage[compress]{cite}
\usepackage{color}
\usepackage{multirow}
\usepackage{array}
\usepackage{tabularx}
\usepackage{svg}
\usepackage{threeparttable} 
\usepackage{caption}
\usepackage{array}
\usepackage{xcolor}
\usepackage[switch, pagewise]{lineno}

\begin{document}
\title{Conditional Diffusion Model for Electrical Impedance Tomography}

\author{Duanpeng Shi, Wendong Zheng, Di Guo*, Huaping Liu*
\thanks{Duanpeng Shi and Di Guo is with the School of Artificial Intelligence, Beijing University of Posts and Telecommunications, Beijing, China.}
\thanks{Wendong Zheng is with the School of Electrical Engineering and Automation, Tianjin University of Technology, Tianjin, China}
\thanks{Huaping Liu is with the Department of Computer Science and Technology, Tsinghua University, Beijing, China, and also with Beijing National Research Center for Information Science and Technology, China.}
\thanks{Duanpeng Shi and Wendong Zheng contributed equally to this work.}
\thanks{Corresponding author: Di Guo (guodi.gd@gmail.com), Huaping Liu (hpliu@tsinghua.edu.cn).}%

\thanks{Manuscript accepted on October 29, 2024; waited to be published.}}

\markboth{IEEE TRANSACTIONS ON INSTRUMENTATION AND MEASUREMENT}%
{Shell \MakeLowercase{\textit{et al.}}: A Sample Article Using IEEEtran.cls for IEEE Journals}


\maketitle

\begin{abstract}

Electrical impedance tomography (EIT) is a non-invasive imaging technique, which has been widely used in the fields of industrial inspection, medical monitoring and tactile sensing. However, due to the inherent non-linearity and ill-conditioned nature of the EIT inverse problem, the reconstructed image is highly sensitive to the measured data, and random noise artifacts often appear in the reconstructed image, which greatly limits the application of EIT. 
To address this issue, a conditional diffusion model with voltage consistency (CDMVC) is proposed in this study. The method consists of a pre-imaging module, a conditional diffusion model for reconstruction, a forward voltage constraint network and a scheme of voltage consistency constraint during sampling process. The pre-imaging module is employed to generate the initial reconstruction. This serves as a condition for training the conditional diffusion model. Finally, based on the forward voltage constraint network, a voltage consistency constraint is implemented in the sampling phase to incorporate forward information of EIT, thereby enhancing imaging quality. A more complete dataset, including both common and complex concave shapes, is generated. The proposed method is validated using both simulation and physical experiments. Experimental results demonstrate that our method can significantly improves the quality of reconstructed images. In addition, experimental results also demonstrate that our method has good robustness and generalization performance.

\end{abstract}


\section{Introduction}
\IEEEPARstart{E}{lectrical} impedance tomography (EIT) is an imaging technology that uses current injection and voltage measurement to create an image of the internal impedance distribution. Through the analyzed images, functional and structural information about the objects of interest can be generated. Due to its non-invasive nature, lack of radiation, low power consumption, and rapid response \cite{miao2014roi}, EIT is widely applied in medical imaging \cite{murphy2016absolute}, industrial non-destructive testing \cite{wei2016super,yao2017application}, robotic sensing\cite{zhou2024bioinspired} and biomedical research \cite{xu2018development}. However, EIT faces several issues and challenges in practical applications:
Firstly, compared to computed tomography (CT) and magnetic resonance imaging (MRI), EIT has relatively low spatial resolution, which limits its use in medical and industrial applications requiring high-resolution imaging. Secondly, EIT measurements are susceptible to noise and artifacts, which can compromise the accuracy of image reconstruction. Thirdly, the image reconstruction algorithms of EIT involve solving an inverse problem that is inherently nonlinear and ill-posed \cite{gamio2003interpretation}. Noisy data can severely degrade the quality of reconstructed images, leading to inaccuracies in interpretation and analysis. This challenge is compounded by factors such as sensor limitations, measurement noise, and environmental interference. These factors place stringent demands on the robustness and noise resilience of EIT imaging methods. While various approaches have been proposed to mitigate these effects, the quality of the resulting images often fails to meet the standards required for practical applications, particularly in high-noise and complex environments. These limitations continue to hinder further advancements in the accuracy and reliability of EIT imaging systems.

Traditional methods have been widely proposed to address EIT image reconstruction. To handle non-linearity, researchers have proposed one-step linearization \cite{cheney1990noser,jin2008augmented}, iterative linearization \cite{murphy2016absolute}, and direct nonlinear methods \cite{alsaker2017direct}. To address ill-posedness, regularization methods such as Tikhonov regularization \cite{vauhkonen1998tikhonov}, Total Variation (TV) \cite{jung2014impedance}, and compressed sensing \cite{gehre2012sparsity} are commonly used. Yet, these methods struggle with selecting appropriate regularization terms and lack of prior knowledge, which impedes improvements in image quality.

With advancements in computing power and the rise of big data, deep learning has made significant breakthroughs in various fields. Several neural networks have been developed for EIT image reconstruction, including LeNet-based convolutional neural network (CNN) \cite{tan2018image}, V-shaped dense denoising networks (VDD-Net) \cite{zhang2022v}, densely connected convolutional neural networks \cite{li2020electrical}, and multilayer autoencoders (MLAE) \cite{chen2021deep}. Unlike traditional approaches that focus on regularized minimization problems in EIT, deep learning leverages its ability to map complex nonlinear representations between data distributions.

Given that the core of the EIT inverse problem can be viewed as an image generation problem under specific conditions, the rapidly advancing deep generative models (DGMs) show potential for solving EIT-related issues. In recent years, DGMs have flourished and garnered extensive research attention in artificial intelligence \cite{kingma2019introduction,goodfellow2020generative,rezende2015variational,ho2020denoising,song2020score}. Inspired by DGMs, researchers have explored various techniques for EIT reconstruction, such as generative adversarial network (GAN) enhanced methods \cite{chen2024mitnet,chen2020electrical,zhang2022image}. These approaches aim to enhance initial reconstructions produced by traditional algorithms. Compared to CNN-based methods, GAN-based techniques often preserve sharper details. However, GAN-based algorithms face challenges such as the saddle point problem, difficulties in training, and a lack of diversity in generated images \cite{zhang2022image}. Moreover, diffusion models have demonstrated impressive results in generation tasks \cite{ho2020denoising,song2020score} and have been applied to the EIT reconstruction problem \cite{zhou2024deep}.

In order to address the aforementioned issues and maximize the utility of the measured voltage, we propose a conditional diffusion model with voltage consistency for EIT. Specifically, the prior knowledge of the forward process is integrated into the conditional diffusion model and introduced during the sampling phase to enhance imaging performance. Our contributions are summarized as follows

\begin{enumerate}[label=\arabic*),leftmargin=2em]

    \item A novel conditional diffusion model is proposed for solving the EIT inverse imaging problem. In particular, using the initial conductivity image reconstructed from traditional reconstruction methods as a condition, the diffusion model is used to effectively improve EIT reconstruction performance and generates high-resolution conductivity image.

    \item In order to incorporate the prior knowledge of the forward process into the conditional diffusion model, a forward constraint network is proposed. Specifically, a lightweight network is designed to establish the relationship between conductivity distribution and boundary voltage, which serves as a kind of physical constraint to further enhance the generation performance of the diffusion model.
    
   \item A comprehensive dataset is constructed for EIT imaging research, and the effectiveness of the proposed method is verified on this dataset. Experimental results demonstrate that the proposed method exhibits robust and accurate reconstruction performance. Furthermore, the method is applied to an actual EIT sensor to further validate its effectiveness.

\end{enumerate}
The structure of the paper is organized as follows.
Section \ref{related work} reviews the existing literature and previous studies.
Section \ref{background} introduces the forward and inverse problems in the context of EIT.
Section \ref{method} details the proposed approach, including the theoretical foundation, algorithmic design, and implementation specifics.
Section \ref{dataset} describes the generation of the dataset. The simulation and physical experiments are presented in Section \ref{simulaton experiment} and \ref{physical experiment}.

And Section \ref{conclusion} comes to the conclusion.

\section{Related work}
\label{related work}
\subsection{Deep Network Model for Inverse Imaging}

The development of deep learning has provided an effective method for inverse imaging. With its capability to learn the nonlinear mapping between input and output, deep learning can avoid the time-consuming process of solving the forward problem. Consequently, deep learning methods are widely applied to solving the EIT problem.

A CNN structure based on LeNet is proposed in \cite{tan2018image}, where dropout layers and moving averages are added to enhance the model's robustness. This approach marks a significant step towards applying deep learning to EIT problems. To handle more complex situations, more sophisticated deep neural networks have been designed. The V-shaped dense denoising net (VDD-Net) consists of an "encoder–decoder" framework for feature extraction and a dense CNN module for filtering reconstruction artifacts \cite{zhang2022v}. Building on the V-shaped network, channel attention and coordinate attention mechanisms have been developed to construct electrical parameter distributions and learn spatial distribution information, respectively \cite{wang2023electrical}. To better mitigate the issue of vanishing gradients in deep network structures, a multilayer autoencoder (MLAE) has been proposed \cite{chen2021deep}. It utilizes a strategy of layer-wise self-supervised and supervised fine-tuning to optimize parameters. Additionally, sparse regularization terms and ${L_2}$
  regularization factors are introduced to avoid overfitting during network training.

Although deep learning has significantly improved imaging quality compared to traditional algorithms, it still faces challenges such as the low-dimensional voltage data not providing sufficient reconstruction information and the risk of losing sharp details.
\subsection{Diffusion-Based Generative Models}

Diffusion models are a family of probabilistic generative models that define the generation process as the reverse process of adding noise. 
In contrast to traditional deep learning models like CNNs or autoencoders, diffusion models are generative and can model the underlying data distribution, offering enhanced robustness against noisy or incomplete EIT measurements. Additionally, they enable progressive refinements for more accurate and controlled outputs.

They are primarily based on three predominant formulations: denoising diffusion probabilistic models (DDPMs) \cite{ho2020denoising}, score-based generative models (SGMs) \cite{song2019generative}, and stochastic differential equations (SDE) \cite{song2020score}. 
To simplify and improve the backward processes in neural networks, methods such as Stable Diffusion \cite{rombach2022high} and DVDP \cite{zhang2023dimensionality} have explored training diffusion models in a learned latent space. Several approaches have been proposed to address challenges in bridging arbitrary distributions, particularly in tasks like image-to-image translation \cite{heitz2023iterative}. For example, Rectified Flow \cite{liu2023flow} introduces additional steps to straighten the bridging process.
Sampling from diffusion models is a crucial step in generating samples, typically requiring iterative methods involving numerous evaluation steps. To accelerate the sampling process while enhancing sample quality, fast sampling methods such as DPM-solver \cite{lu2022dpm} and DDIM \cite{song2020denoising} have been proposed. Moreover, to guide the generation directions of diffusion models, conditional diffusion models have been developed, including classifier-free guidance \cite{ho2022classifier} and classifier guidance \cite{dhariwal2021diffusion}. Conditioning mechanisms generally include four types: concatenation, gradient-based, cross-attention, and adaptive layer normalization (AdaLN).

Due to their flexibility and strength, diffusion models are widely used to address a variety of challenging real-world tasks, such as computer vision \cite{li2022srdiff}, natural language processing \cite{austin2021structured}, and temporal data modeling \cite{tashiro2021csdi}. {To the best of our knowledge, the CSD$^{*}$ method is the first to incorporate diffusion models into the EIT domain, treating score-based diffusion as a post-processing operator following the Gauss-Newton method \cite{zhou2024deep}. However, it employed an excessive number of sampling steps.} Nonetheless, it struggles with handling complex data distributions due to the absence of guidance information during training and lacks hard constraints on measured voltage. Inspired by the above-mentioned works, we utilize DDPM as the base framework to obtain the prior for different conductivity distributions, with initial reconstruction serving as the condition. The concatenation method is employed to impart this condition to the framework.

\section{Preliminaries} 
\label{background}

EIT consists of a forward problem and an inverse problem. The forward problem of Electrical Impedance Tomography (EIT) involves calculating the voltage distribution across electrodes given the conductivity distribution and injected current. In contrast, the inverse problem of EIT aims to estimate the internal conductivity distribution based on the observed voltage distribution across the electrodes.

{
\textbf{\textit{EIT forward problem:}}
The time-difference EIT is utilized to reconstructs the changes in conductivity between $V$ ($V=v_{2}-v_{1}$) and $\sigma = {\sigma _2} - {\sigma _1}$. $v_{1}$ represents the reference voltage measurement that is resumed as noise free and $v_{2}$ denotes the voltage measurement with inclusion within the measured domain. $\sigma _2$ is the current conductivity distribution, and  $\sigma _1$ is the conductivity distribution during the reference measurement. 
The forward problem can be formulated as 
\begin{equation}
\label{FEM}
   V = y(\sigma, I),
\end{equation}
where $I$ denotes the excitation current and the $y(\sigma, I)$ represents the forward model that map $\sigma$ and $I$ to $V$. 
}

{
\textbf{\textit{EIT inverse problem:}} 
Thanks to the rapid advancements in deep learning, significant progress has been made in deep learning-based image reconstruction methods. The mathematical model for EIT image reconstruction using deep learning can be formulated as follows:
\begin{equation}
\label{eq:inverse_model}
    {\widehat \sigma ^*} = \mathop {\arg \min }\limits_{\widehat \sigma } \frac{1}{2}(\underbrace {\left\| {{\sigma _{true}} - \widehat \sigma } \right\|_2^2}_{{L_{\text{data}}}} + \underbrace {\left\| {V - P(\widehat \sigma )} \right\|_2^2}_{{L_{\text{physics}}}}),
\end{equation}
where $L_{\text{data}}$ is the data consistency term, ensuring that the estimated conductivity image $\widehat{\sigma}$ is close to the ground truth $\sigma_{\text{true}}$. The estimated conductivity $\widehat{\sigma}$ is typically obtained through a convolutional neural network (CNN), $F_\theta(\cdot)$, where $\theta$ represents the learnable parameters. The $L_{\text{physics}}$ term enforces physical consistency by ensuring that the conductivity distribution output by the network $\widehat{\sigma}$ is physically plausible. Specifically, $P$ denotes the forward model and V is the measured voltage.
}

\begin{figure*}[htpb]
	\centering
	\includegraphics[width=\textwidth]{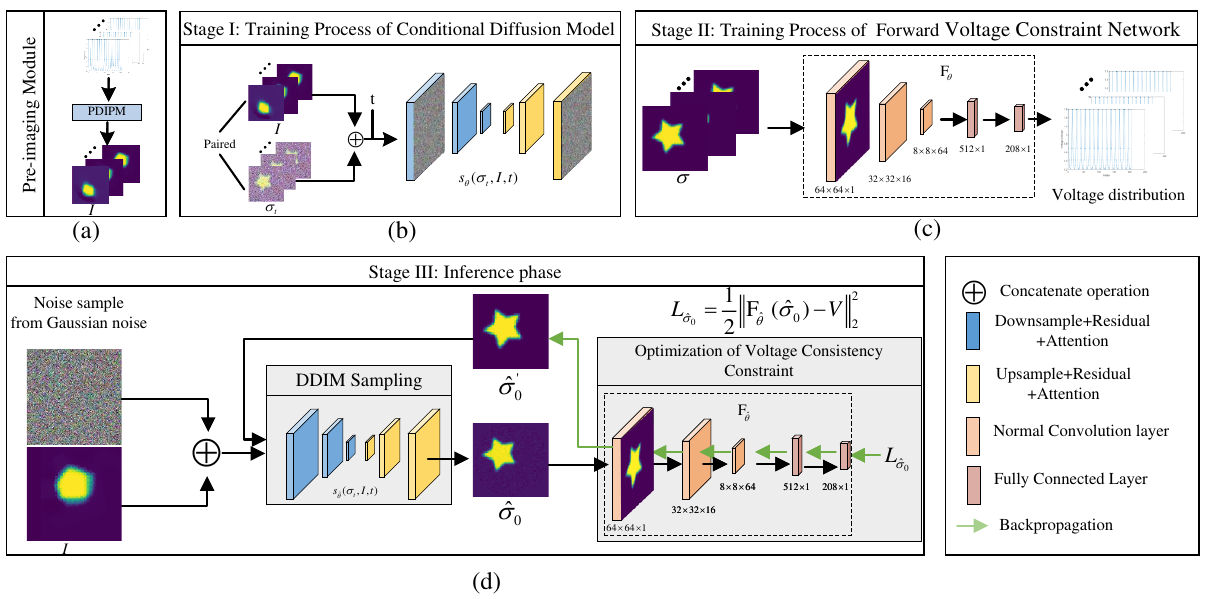}
	\caption{It includes four parts. As for inference phase, the input consists of the initially reconstructed conductivity concatenated with a noise sample from Gaussian noise. This concatenated feature is fed into the DDIM sampling process, where voltage consistency constraint optimization is applied to outputs of the intermediate steps.}
	\label{Framework}
\end{figure*}
\section{METHOD}
\label{method}

In this paper, the EIT inverse problem is treated as an image reconstruction task. We take advantage of conditional diffusion model in image generation. The initial reconstruction is utilized to guide the generation process and the voltage consistency constraint is implement on the output of sampling steps to improve image quality. Specifically, the inference phase of our proposed method is illustrated in Fig. \ref{Framework}(d). Paired images, including an initially constructed image $I$ and an image-sized noise sample from Gaussian noise, are input into the trained diffusion model in a concatenated manner. A fast sampling method, DDIM sampling, is adopted to generate the conductivity distribution. The outputs of certain sampling steps are then optimized with a designed voltage consistency constraint to refine the generated conductivity. This section introduces our proposed method in the following four parts.

\subsection{Pre-imaging Module}
This subsection discusses how to construct the training samples. The Gauss-Newton-related methods are commonly used for the initial pre-reconstruction of the conductivity distribution \cite{chen2020electrical, zhou2024deep}. However, these methods are sensitive to outliers and result in smooth spatial distributions. To improve the quality of the initial reconstructed image, we utilize the PDIPM with total variation as the prior \cite{borsic2012primal}, as shown in the Fig. \ref{Framework}(a). 
{
Total Variation (TV) regularization is a technique used to handle noise and enhance edge information. The objective function can be written as:
\begin{equation}
\min _\sigma \frac{1}{2}\|J \sigma-b\|_2^2+\lambda \cdot \operatorname{TV}(\sigma)
\end{equation}
where $J \sigma-b$ is the data term, representing the difference between the measured data and the forward model. $\lambda$ is the regularization parameter that balances the data term and the regularization term and $ \operatorname{TV}(\sigma)$ is the TV regularization term, which preserves the edge information in the image. TV regularization is nonlinear and non-smooth because it involves the L1-norm of the image gradient. This makes it difficult for traditional optimization algorithms to be applied directly. The Interior Point Method (PDIPM) is used to handle this non-smooth problem by optimizing both the primal and dual problems simultaneously. Firstly, the barrier function is introduced to ensure that each iteration stays within the feasible region:
\begin{equation}
    \mathop {\min }\limits_\sigma  \frac{1}{2}\left\| {J\sigma  - b} \right\|_2^2 + \lambda  \cdot {\mathop{\rm TV}\nolimits} (\sigma ) - \mu \sum\limits_i {\log ({\sigma _i})}    
\end{equation}
The $\log ({\sigma _i})$ barrier function ensures that $\sigma $ remains positive. Then the Karush-Kuhn-Tucker (KKT) conditions \cite{bertsekas1997nonlinear} is constructed, which represent the balance between the primal and dual problems:
\begin{equation}
    \nabla L(\sigma ,\lambda ) = 0  
\end{equation}
$ L(\sigma ,\lambda )$  is the primal-dual Lagrangian function. The KKT conditions describe the optimal solution that satisfies both the data term and the constraints. Finally, The interior point method uses Newton's method to iteratively solve the KKT equations. Specifically, it update the primal variable $\sigma$, dual variable $\lambda$ and $\mu$ sequentially. 
The PDIPM algorithm finally outputs a solution $I$ that has been regularized using TV, preserving edge information in the image while suppressing noise.
}

\subsection{Conditional Diffusion Model Based on Initial Reconstruction}
{
In this section, we provide a comprehensive description of how the conditional diffusion model is trained using the initial reconstructions, denoted by $I$. Fig. \ref{Framework}(b) illustrates our approach, where the initially reconstructed conductivity $I$ serves as a condition to guide the image generation process. The underlying architecture of our model is based on a U-Net, which is used to predict the noise introduced during the forward diffusion process.
}

{
The forward process in our conditional diffusion model is modeled by the variance-preserving (VP) form of the Stochastic Differential Equation (SDE) \cite{song2020score}, which gradually corrupts the true conductivity distribution by adding noise at each time step. This process is particularly suitable for EIT because it introduces a controlled degradation, making it easier for the reverse process to refine details and recover fine structures in the reconstructed image. The VP-SDE governing the forward diffusion process is given by:
}
{
\begin{equation} \label{VP_form} d\sigma _t = - \frac{1}{2}\beta (t)\sigma _t dt + \sqrt {\beta (t)} dw, \end{equation}
}

{
where $\beta (t)$ represents the noise schedule, which controls how noise is added over time. In our implementation, we use a cosine noise schedule, which has been shown to maintain better balance between early and late diffusion steps, making it effective for capturing intricate structures in EIT. Here, $\sigma_t$ represents the noisy version of the true conductivity at time $t$, evolving under the influence of the Wiener process $w$.
}

{
During the reverse process, our goal is to reconstruct the true conductivity distribution starting from the noisy data. This reverse-time SDE allows us to undo the corruption introduced by the forward diffusion process, with the initially reconstructed conductivity $I$ acting as a condition that guides this refinement. The reverse SDE is formulated as:
}
{
\begin{equation} \label{reverse_process} d\sigma _t = \left[ - \frac{{\beta (t)}}{2}\sigma _t - \beta (t){\nabla _{{\sigma _t}}}\log {p_t}({\sigma _t}|I)\right] dt + \sqrt {\beta (t)} d\bar w, \end{equation}
}

{
where ${p_t}({\sigma_t}|I)$ represents the conditional distribution of the noisy conductivity $\sigma_t$ given the initial reconstruction $I$, and $\nabla{\sigma_t} \log p_t(\sigma_t|I)$ is the gradient of the log-probability (i.e., the score function) at time $t$. This score function is essential for guiding the reverse process toward more accurate reconstructions. The Wiener process $d\bar{w}$ models the stochastic noise in this reverse trajectory. The role of the initial reconstruction $I$ is critical in this step: it encodes coarse structural information such as location, approximate shape, and conductivity values of inclusions. This conditioning helps narrow down the search space, making the reverse process more efficient and accurate. As a result, the reverse process can focus on recovering fine details that are challenging to obtain using conventional reconstruction techniques in EIT.
}

{
The neural network $s_\theta(\sigma_t, I, t)$ is trained to predict the noise added to the conductivity during the forward diffusion process. This is achieved using denoising score matching, where the model learns to approximate the gradient of the log-likelihood (Stein score function) with respect to the noisy conductivity:
}
{
\begin{equation} 
\label{training diffusion}
    \mathop {\min }\limits_\theta \mathbb{E}\left[\parallel {s_\theta }({\sigma _t},I,t) - {\nabla _{{\sigma _t}}}\log p({\sigma _t}|I)\parallel _2^2\right],
\end{equation}
}

{
where, $t$ is uniformly sampled from $[0,T]$, and $\sigma_t \sim p(\sigma_t|I)$ is the noisy version of the true conductivity, while $\sigma_0 \sim p_{\text{data}}(\sigma)$ is the ground truth conductivity. The expectation is taken over the entire diffusion process, ensuring that the model learns to denoise samples at all stages of noise corruption.
}

{
The training process aims to minimize the difference between the predicted noise and the actual noise introduced during the forward process. By learning this score function, the model becomes capable of generating samples that follow the reverse diffusion process described in Eq. \ref{reverse_process}, thereby producing accurate conductivity.
Our proposed method builds upon the VP-SDE model (DDPM) \cite{ho2020denoising}, which has shown success in image generation tasks. In the context of EIT, the conditional diffusion model is particularly well-suited for handling the inherent noise and ill-posed nature of the inverse problem. By leveraging the initial reconstruction $I$, which provides limited but valuable information about the location, shape, and approximate value of inclusions, we guide the diffusion process to focus on plausible solutions.
}

{
Specifically, in the forward diffusion process, we start with the ground truth conductivity $\sigma$ and corrupt it with Gaussian noise to generate noisy samples $\sigma_t$. These noisy samples are concatenated with the initial reconstruction $I$ and fed into the U-Net-based model, which predicts the noise component to be removed during the reverse process. This conditioning helps the model to adapt to the EIT-specific reconstruction challenges, such as irregular inclusion shapes and varying noise levels.
}

\subsection{Forward Voltage Constraint Network}

The initially construction based diffusion model is capable of generating the result that close to the real conductivity distribution based on diffusion model,  However, there are no guarantees that the sample explains the data $V$, i.e. ${\rm{F}}({\sigma }) \approx V$, $\sigma $ is the conductivity contribution. In order to fulfill this voltage consistency constraint and further improve the construction quality, we model it as follows:
\begin{equation}
\label{eq:mse constraint}
{L_{MSE}} = \frac{1}{N}\sum\limits_{i = 1}^N {\left\| {{{\rm{F}}_\theta }({\sigma ^i}) - {V^i}} \right\|_2^2},
\end{equation}
where $N$ is the number of training samples, ${\rm{F}}( \cdot )$ is the nonlinear function that represents forward voltage constraint network (FVCN) with $ \theta $ is the learnable parameter. ${\sigma ^i}$ is true conductivity distribution. $V$ is the true voltage vector and the predicted voltage vector. ${\rm{F}}( \cdot )$ is the forward model which is often implemented as finite element method (FEM). However, the FEM cannot different the conductivity distribution generated from diffusion model. In order to handle this problem, we design the FVCN shown in Fig. \ref{Framework}(c). It is designed relatively simple to avoid massive calculation, which consists of three convolutional layers and two fully connected layers. The mean square error (MSE) loss is exploited for FVCN optimization in the training. After the training process, the 
${{\rm{F}}_{\hat \theta {\rm{ }}}}$ is obtained which is utilized in inference phase. {The forward voltage constraint network is designed to enforce voltage consistency by ensuring that the predicted voltage values during the sampling process align with the measured voltages from the EIT system.}
\begin{algorithm}
\caption{Sampling Algorithm with Voltage Consistency Constraint during Sampling Process}
\label{alg:during_sample_algorithm}
\begin{algorithmic}[1] 
\Require $N$, $I$, ${{\rm{F}}_{\hat{\theta}}}$, $V$, $\eta _t$
\Ensure The optimized sampling result $\widehat{\sigma}_0$
\State ${\bf{x}}_N \sim \mathcal{N}({\bf{0}}, {\bf{I}})$
\For {$t = N$ to $1$}
    \State ${\bf{z}} \sim \mathcal{N}({\bf{0}}, {\bf{I}})$
    \State $\widehat{\bf{s}} \leftarrow {\bf{s}}_\theta({\sigma}_t, I, t)$
    \State $\widehat{\sigma}_0 \leftarrow \frac{1}{\sqrt{\bar{\alpha}_t}} ({\sigma}_t + (1 - \bar{\alpha}_t) \widehat{\bf{s}})$
    \State $\sigma_{t - 1}' \leftarrow \sqrt{{\bar{\alpha}}_{t - 1}} \widehat{\sigma}_0 + \sqrt{1 - {\bar{\alpha}}_{t - 1} - \tilde{\sigma}_t^2} \, \widehat{\bf{s}} + \eta_t \, {\bf{z}}$

    \If {$t \% 10 == 0$} \label{line:C}
       \State ${\hat \sigma _0}^\prime {\rm{ = arg}}{\mkern 1mu} \mathop {{\rm{min}}}\limits_{{{\hat \sigma }_0}} {\mkern 1mu} \frac{1}{2}\left\| {{{\rm{F}}_{\hat \theta }}({{\hat \sigma }_0}) - V} \right\|_2^2$
       \State ${\sigma _{t - 1}} \leftarrow {{\hat \sigma }_0}^\prime $
    \Else
        \State ${\sigma_{t - 1}} \leftarrow \sigma_{t - 1}'$
    \EndIf
\EndFor
\State \Return $\widehat{\sigma}_0$
\end{algorithmic}
\end{algorithm}

\subsection{Inference Phase}
\label{subsec:inference}
The inference phase involves generate the samples based on the initially reconstructed conductivity $I$, and conduct the voltage consistency constraint is conducted based on trained FVCN. It is noted that the voltage consistency constraint can function as the plug and play scheme during or after sampling process, improving the reconstruction quality significantly.
To accelerate the speed of sampling, the denoising diffusion implicit model (DDIM) \cite{song2020denoising} is utilized. 

Compared with slow sampling process in \cite{ho2020denoising, song2022pseudoinverse}, DDIMs sampling method greatly improve the sampling speed by defining the diffusion process a non-Markovian process\cite{song2020denoising}. 
We utilize an approximation of the posterior mean as the output of DDIM samping result, as described by the following formula:
\begin{equation}
\label{eq:postior}
{\widehat \sigma _0} = \frac{1}{{\sqrt {{{\bar \alpha }_i}} }}\left( {{\sigma _t} + (1 - {{\bar \alpha }_t}){{\bf{s}}_\theta }\left( {\sigma _t,I,t} \right)} \right),
\end{equation}
$\widehat \sigma _0$ represents the estimate of ground-truth latent vector $\sigma _0$ based on the sample $\sigma _t$.Then, the sample can be updated with the sampling steps as follows: 
\begin{equation}
\label{eq:DDIM}
{\sigma _{t - 1}} = \sqrt {{{\bar \alpha }_{t - 1}}} {\widehat \sigma _0} + \sqrt {1 - {{\bar \alpha }_{t - 1}} - \tilde \sigma _i^2} {s_\theta }\left( {{\sigma _t},I,t} \right) + \eta _t\bf{z},
\end{equation}
where $\alpha _t = 1- \beta (t)$, $\bar{\alpha}_t = \prod_{i=1}^{t} \alpha_t$, $\bf{z} \sim \mathcal{N}(0, I)$, 
$\eta _t=\frac{1-\bar{\alpha}_{t-1}}{1-\bar{\alpha}_t} \beta_t$. 
Based on the output of DDIM sampling steps, we propose to solve an optimization problem on some time steps $t$:
\begin{equation}
\label{eq:voltage-constraint}
{\widehat \sigma _0}'  = \arg \mathop {\min }\limits_{\widehat \sigma _0}  \frac{1}{2}\left\| {v - {{\rm{F}}_{\hat \theta }}({\widehat \sigma _0})} \right\|_2^2,
\end{equation}
which is a hard data consistency that make the generated conductivity consistent with the measurement $v$. ${{\rm{F}}_{\hat \theta {\rm{ }}}}$ is the trained FVCN which functions as a forward model that maps the reconstruction image to the voltage. $\widehat \sigma _0$ is the sample result from DDIM that can be optimized by voltage consistency constraint. This optimization problem can be efficiently solved using iterative methods like gradient descent.
 {Specifically, two steps are taken to refine the conductivity distribution over a set number of sampling steps. First, the Forward Voltage Constraint Network (FVCN) is used to predict the voltage based on the conductivity image generated by the conditional diffusion model. Then, the loss function in Equation (\ref{eq:voltage-constraint}) is applied between the measured and predicted voltages from the FVCN, which is used to optimize the conductivity image produced by the conditional diffusion model.}

To be noted, the voltage consistency constraint can be implemented during or after the sampling process. Algorithm\ref{alg:during_sample_algorithm} shows how to conduct voltage consistency constraint during the sampling process. It is also unnecessary because the beginning sampling would deviate significantly from the real conductivity distribution. Besides, the optimization of voltage consistency constraint is costly if it is implemented for every sampling step. Thus, we devise a skipped-step mechanism for performing voltage consistency constraint on every 10 (or so) iterations of $t$. In Section \ref{subsec:ablation result}, the constraint is complemented after the sampling ends. The experiment results show that both schemes can improve the image quality.

\section{Dataset Construction}
\label{dataset}
To demonstrate the effectiveness of the proposed reconstruction method, we construct a large-scale dataset. It is challenging to obtain a large amount of conductivity distribution data and corresponding boundary voltages in actual experiments. Therefore, we use the EIDORS tools in MATLAB to generate samples for model training. A 16-electrode model is applied as the simulation model, which aligns with the physical experimental setup. The adjacent excitation–adjacent measurement strategy is adopted. Specifically, 16 pairs of adjacent electrodes are sequentially excited with a 0.01A current. The finite-element method is employed to solve the steady-state forward problem of EIT, utilizing adaptive triangular meshes available in the mesh generation toolbar of EIDORS. 208 measurements are acquired, constituting a complete set of cross-sectional test data for one frame \cite{cheney1999electrical}. 

\begin{figure*}[ht]
	\centering
	\includegraphics[width=\textwidth]{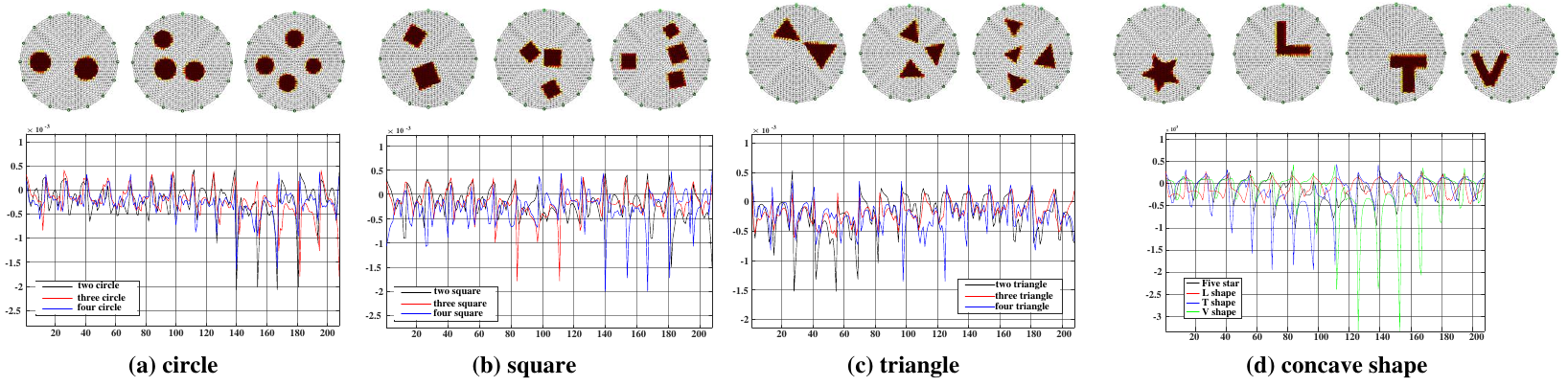}
	\caption{Voltage measurements under different settings}
	\label{all_data}
\end{figure*}
\begin{figure}[ht]
	\centering
	\includegraphics[width=\columnwidth]{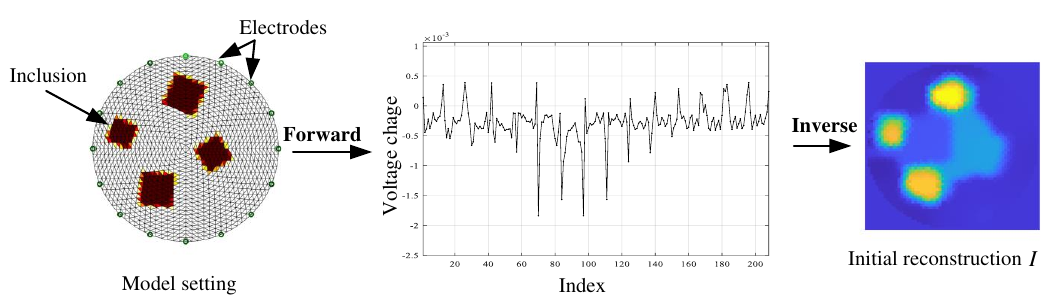}
	\caption{The process of generating training data}
	\label{data}
\end{figure}

The detailed settings of generating the simulation database are as follows. The background conductivity is defined as 0.6 S/m, while inclusions are characterized by a conductivity of 0.003 S/m. As shown in Fig. \ref{data}, the inclusions are set and input into the forward model, which outputs the voltage distribution $U$. Then, the initial reconstruction $I$ is obtained using traditional algorithm PDIPM. Each sample's information consists of three data vectors: the true conductivity distribution $\sigma$, the initially reconstructed conductivity distribution $I$, and the boundary voltage vector $U$. To be noted, $U$ is obtained by solving the forward problem of EIT and represents the difference between the boundary voltage measured with inclusions in the field and that measured without inclusions. To prevent inverse crime, where theoretical components closely related to those used to synthesize and invert data in an inverse problem are the same or very similar, we transform the initial finite element data into 64 × 64 pixel data using inverse distance weighted interpolation.

Based on the aforementioned theoretical foundation, we generate a large-scale dataset, including both common and complex shapes. The common shapes refer to those frequently used in literature, such as circles, triangles, and squares. Notably, the positions and offset angles of the triangles and squares are generated randomly. The complex shapes include concave forms like five-pointed stars, L-shape, T-shape, and V-shape, which have sharp concave corners that are challenging to reconstruct. All types of model settings and their corresponding voltage distributions are shown in Fig. \ref{all_data}. In the simulations, 10,000 samples are generated for each setting, including circles, triangles, and squares with 2, 3, or 4 inclusions respectively, and the aforementioned concave shapes with only one inclusion. 
The randomly shuffled dataset is split into a training set consisting of 9900*13 samples and a testing set of 100*13 samples. Both sets are acquired within the same simulation environment and are independent and identically distributed. 

\section{EXPERIMENT on Simulation DATA}
\label{simulaton experiment}
The proposed conditional diffusion method, which incorporates a voltage consistency constraint, is evaluated through a series of numerical tests, with the corresponding visualization results presented.

\subsection{Evaluation Metrics}
\label{metrics}
{To ensure a consistent comparison with traditional methods and other state-of-the-art techniques, we employ several quantitative metrics to evaluate reconstruction performance, including relative error (RE), structural similarity (SSIM), peak signal-to-noise ratio (PSNR), dynamic range (DR), Mean Squared Error (MSE) and Correlation Coefficient (ICC).}
\begin{itemize}
    \item RE measures the relative difference between the reconstructed and ground truth conductivity distributions:
    \begin{equation}
    \label{eq:RE}
        \text{RE}=\frac{\left\|\sigma^{\prime}-\sigma_{G T}\right\|_1}{\left\|\sigma_{G T}\right\|_1},
    \end{equation}
    where ${{\sigma ^\prime }}$ is the reconstructed conductivity and ${{\sigma _{GT}}}$ is the ground truth conductivity. ${\left\|  \cdot  \right\|_1}$ is defined as the sum of the absolute values of the vector's components.
\end{itemize}
\begin{itemize}
    \item SSIM evaluates the similarity between two images, considering structural information, luminance, and contrast:
    \begin{equation}
    \label{eq:SSIM}
        \text{SSIM}=\frac{4 \cdot \bar{\sigma} \cdot \overline{\hat{\sigma}} \cdot \operatorname{Cov}(\sigma, \hat{\sigma})}{\left(\bar{\sigma}^2+\overline{\hat{\sigma}}^2\right) \cdot\left(\operatorname{var}(\sigma)^2+\operatorname{var}(\hat{\sigma})^2\right)},
    \end{equation}
    where ${\bar \sigma }$ and ${\bar \hat \sigma }$ are the mean values of the original and reconstructed images. ${{\mathop{\rm Cov}\nolimits} (\sigma ,\hat \sigma )}$ is the covariance,  ${{\mathop{\rm var}} (\sigma )}$ and ${{\mathop{\rm var}} (\hat \sigma )} $are the variances of the original and reconstructed images.
   
\end{itemize}
\begin{figure*}[htpb]
    \centering
    \includegraphics[width=\textwidth]{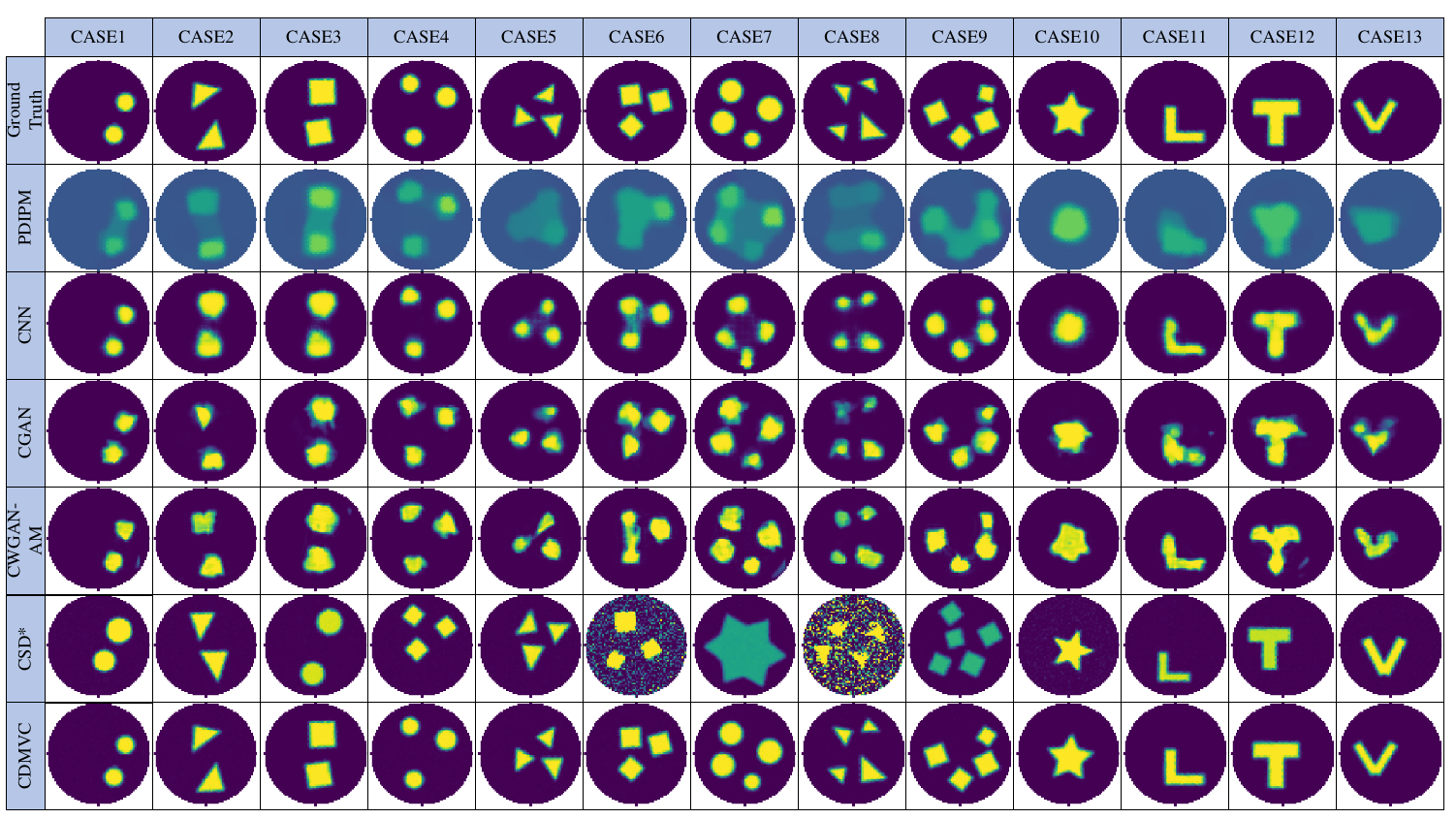}
    \caption{Visual results of different algorithms}
    \label{comparative}
\end{figure*}
\begin{itemize}
    \item PSNR quantifies the quality of reconstructed images:
    \begin{equation}
    \label{eq:PNSR}
        \text{PSNR}=10 \cdot \log _{10}\left(\frac{M A X_I^2}{M S E}\right).
    \end{equation}
    $MA{X_I}$ is the maximum possible pixel value of the image. 
\end{itemize}
\begin{itemize}
    \item DR measures the ratio of the range of reconstructed conductivity values to the ground truth:
    \begin{equation}
    \label{eq:DR}
        \text{DR}=\frac{\max \left(\sigma^{\prime}\right)-\min \left(\sigma^{\prime}\right)}{\max \left(\sigma_{G T}\right)-\min \left(\sigma_{G T}\right)} \times 100 \%.
    \end{equation}
    It is noted that DR suggests better results when it approaches 1.
\end{itemize}
\begin{itemize}
    \item {The Correlation Coefficient (CC) quantifies the degree to which two variables are linearly related:
\begin{equation}
    \label{eq:CC}
        \text{CC}  = \frac{\sum (\sigma_i' - \bar{\sigma})(\sigma_{GT}^i - \bar{\sigma}_{GT})}{\sqrt{\sum (\sigma_i' - \bar{\sigma})^2 \sum (\sigma_{GT}^i - \bar{\sigma}_{GT})^2}}.
\end{equation}
where $\sigma_i'$ and $\sigma_{GT}^i$ represent the pixel values of the reconstructed image and the ground truth image, respectively. }
\end{itemize}

\subsection{Comparative Methods}

\label{comparative methods introduction}
To validate the effectiveness of our method, various of comparative methods are adopted, including numerical methods, CNN-based methods and generative methods. 
\begin{itemize}
    \item \textbf{\textit{PDIPM}}:
    Primal Dual Interior Point Methods (PDIPM) with total variation as the prior is utilized as numerical method. 
    \item \textbf{\textit{CNN-based methods}}: The CNN method refers to the V Dense Net (VD-Net) method \cite{li2020electrical} while it is not completely in accord with it. Because the lower resolution of our data is much lower, the network architecture needs slight adjustments to prevent overfitting. Specifically, the designed architecture consists of an encoder and a decoder. The encoder comprises two convolutional layers with kernel sizes of 5 and padding of 3, each followed by batch normalization, ReLU activation, and max-pooling layers. The decoder then reconstructs the image using two transposed convolutional layers.

    \item \textbf{\textit{CGAN}}: A reconstruction method based on a conditional generative adversarial network is proposed to mitigate the problem of blurred reconstructed images and the lack of detailed features \cite{chen2020electrical}. Compared to the original paper, a 10:1 training scheme for the generator and discriminator is applied to better train the CGAN, which is challenging to converge. The PDIPM algorithm is used to reconstruct the conductivity distribution in accordance with our setup for a fair comparison.
    \item \textbf{\textit{CWGAN-AM}}: The conditional Wasserstein generative adversarial network with attention mechanism (CWGAN-AM) consists of an imaging module, a generator, and a discriminator for EIT image reconstruction \cite{zhang2022image}. We adjusted the number of neurons in the imaging module's connection layer to match the size of our data.
    \item \textbf{\textit{$\boldsymbol{\mathrm{CS}D^*}$}}: This is the first method to introduce the diffusion model into the EIT area. Specifically, it treats the score-based diffusion as a post-processing operator for the solution of the Gauss-Newton method. The sampling employs a speed-up method, which makes it difficult to reconstruct complex shapes due to the poor initial reconstruction.
\end{itemize}

\subsection{Comparison of Different Reconstruction Methods with Simulation Data}
\label{subsec:simulation result}
\begin{figure}[htpb]
	\centering
	\includegraphics[width=\columnwidth]{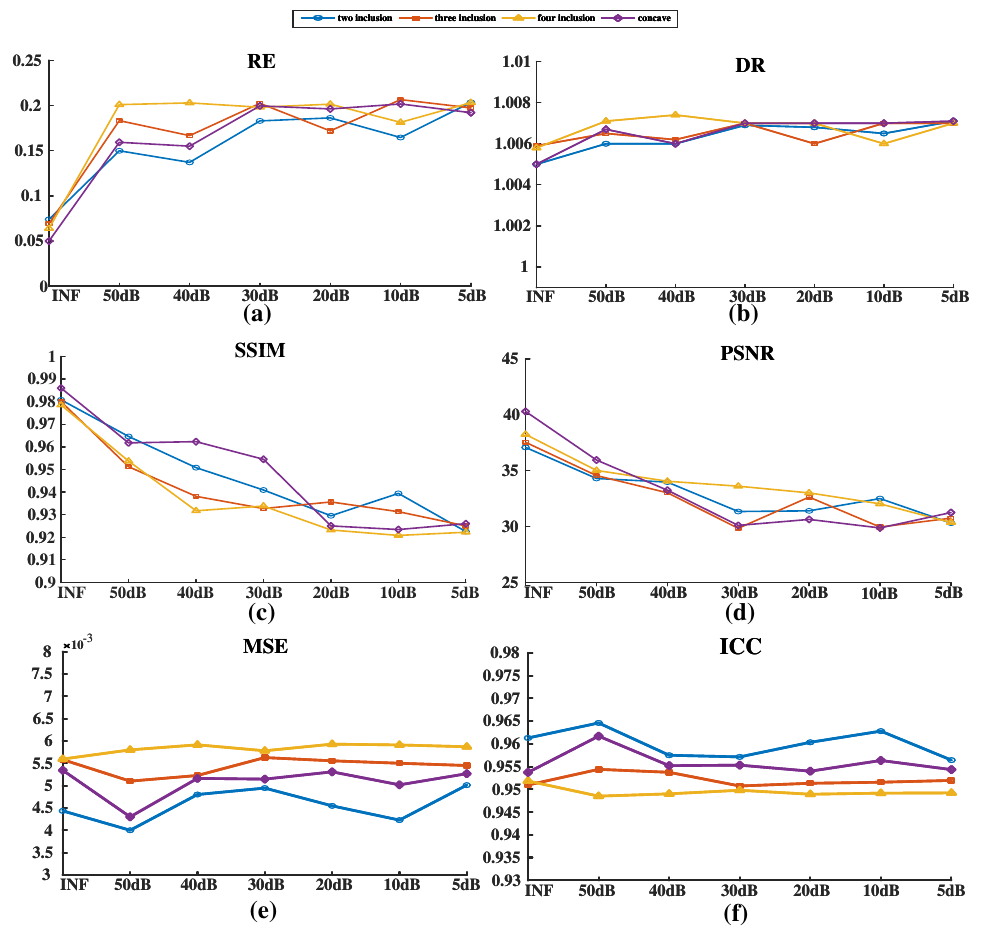}
	\caption{The metrics for different number of inclusions at different noise level. }
	\label{anti-noise status}
\end{figure}

In this section, we compare the performance of our method with various comparative methods introduced in Section \ref{comparative methods introduction}.
Performance metrics of different methods are been compiled in TABLE \ref{tab:algo_comparison}. Our method achieves the highest imaging accuracy among the comparative methods. Specifically, the RE and MSE of our method are 0.0634 and 0.0007 respectively, which are the lowest among all the methods. Additionally, the SSIM and PSNR reach 0.9819 and 38.4498, respectively. Besides, the DR and CC are closest to 1. More detailed statistical data can refer to Fig. \ref{anti-noise status} when the noise is INF. It shows the performance of different number of inclusions. As the number of inclusions increases, the RE increases, while the SSIM and PSNR decrease, and the DR deviates further from 1. This trend indicates that a higher number of inclusions presents greater reconstruction challenges. Notably, concave inclusions achieve the best performance.
\begin{figure*}[htpb]
	\centering
	\includegraphics[width=\textwidth]{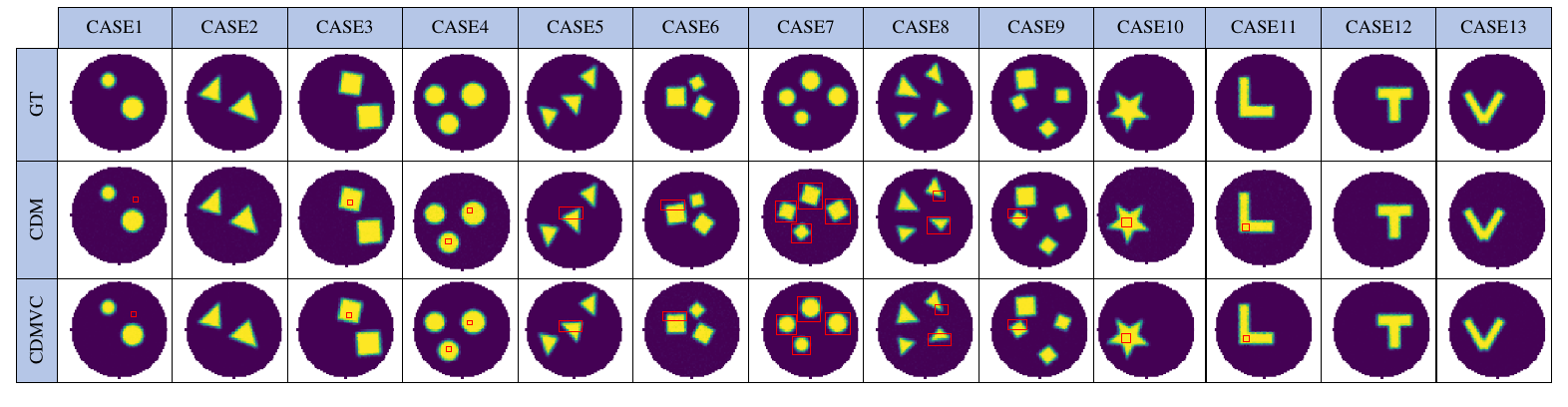}
	\caption{The ablation experiment of our method}
	\label{ablation}
\end{figure*}

\begin{table}[h]
    \centering
    \caption{Comparison of Metrics for Comparative and Our Method}
    \resizebox{\columnwidth}{!}{ 
        \begin{tabular}{lcccccc}
            \toprule
            Algorithm & RE &  SSIM & PSNR & {MSE} & {CC} & DR* \\
            \midrule
            PDIPM  & 2.6209 &  0.4157 & 11.5781 & {0.0699} & {0.6230} & 0.7119 \\
            CNN  & 0.2670 & 0.8886 & 22.4915 & {0.0066} &  {0.9482} & 0.9775 \\
            CGAN  & 0.3923 & 0.8155 & 19.1328 & {0.0130} & {0.8944} & 1.04 \\
            WGAN  & 0.3810 & 0.8300 & 19.2222 & {0.0135} & {0.8988} & 1.318 \\
            ${\rm{CS}}{{\rm{D}}^*}$ & 0.4440 & 0.8177 & 21.7015 & {0.009} & {0.885} & 1.3171 \\
            \textbf{CDMVC}  & \textbf{0.0634} &  \textbf{0.9819} & \textbf{38.4498} & {\textbf{0.0007}} & {\textbf{0.9934}} & \textbf{1.0059} \\
            \bottomrule
        \end{tabular}
    }
    \label{tab:algo_comparison}
    \captionsetup{justification=raggedright,singlelinecheck=false}
    \caption*{* DR values: The closer to 1, the better the performance.}
\end{table}

To visually display the results of each method, Fig. \ref{comparative} shows the reconstruction outcomes for various cases. CASE 1-3 illustrate the reconstructed results for two inclusions of common shapes, CASE 4-10 display the reconstructions for an increasing number of inclusions with common shapes, and CASE 11-13 present the reconstructed results for convex inclusions.

From CASE 1-3 in Fig. \ref{comparative}, it is evident that the PDIPM method can reconstruct the locations of two inclusions relatively accurately but struggles with accurately capturing the corners of triangles or squares. The CNN method improves performance by capturing nonlinear relationships, but it often loses corner information for triangles and squares. Generative algorithms enhance performance in retaining angle information but are still not entirely precise. The ${\rm{CS}}{{\rm{D}}^*}$ method provides clearer images but faces accuracy issues due to the similarity of initial reconstructions, such as those for circles and squares.
As the number of inclusions increases, as shown in CASE 4-10 in Fig. \ref{comparative}, the PDIPM struggles to distinguish shapes and positions. The CNN method performs well for circular inclusions but fails with triangles and squares. Generative methods retain some corner information, as seen in CASE 9, where parts of the square’s angles are reconstructed. However, these reconstructions are not clear and contain many artifacts. The ${\rm{CS}}{{\rm{D}}^*}$ method's reconstructions are unstable due to the lack of guidance in the training process.
CASE 11-13 in Fig. \ref{comparative} show results for convex inclusions. Similarly, PDIPM loses corner information, rendering the five-pointed star shape unrecognizable. The CNN method also fails to accurately capture corners. GAN-based methods improve performance but still produce unclear borders and inaccurate corners, as seen in CASE10. The ${\rm{CS}}{{\rm{D}}^*}$ method reconstructs clearer shapes but with less accurate positions.

Our method effectively handles complex cases. The reconstructed images exhibit clear borders and retain corner information across different cases, from triangles and squares to concave shapes. Although the angles of triangles in CASE5 and CASE8 are not completely accurate, our method achieves the best results among the comparative methods. The conditional diffusion model with voltage consistency constraint has the capability of reconstructing images with the limited information imparted from the initial reconstruction.

\subsection{Analysis of the Effect of Voltage Consistency Constraint}
\label{subsec:ablation result}

In this section, we analyze the effect of voltage consistency module through ablation experiments. The metrics of conditional diffusion model, VC constraint after sampling and VC constraint during sampling are calculated and presented in TABLE \ref{tab:algos}. The VC constraint during sampling yields the best performance, with the lowest RE and MSE, and the highest SSIM, CC, and PSNR values. Additionally, the DR metric is closest to 1. Therefore, we designate it as CDMVC, which is used in subsequent experiments. Therefore, we refer it as CDMVC which is used in the subsequent experiment. To present the reconstruction results more intuitively, the visualization results are further shown in Fig. \ref{ablation}.
\begin{table}[htpb]
    \caption{Ablation of Our Method}
    \resizebox{\columnwidth}{!}{
        \begin{tabular}{lcccccc}
            \toprule
            Algorithm & RE &  SSIM & PSNR & {MSE} & {CC} & DR \\
            \midrule
            CDM  & 0.0931 & 0.9548 & 36.7344 & {0.0008} & {0.9909} & 1.0336 \\
            VC After Sampling   & 0.0671 & 0.9677 & 37.7015 & {0.00075} & {0.9911} & 1.0171 \\
            \textbf{VC During Sampling}  & \textbf{0.0634} & \textbf{0.9819} & \textbf{38.4498} & {\textbf{0.0007}} & {\textbf{0.9934}} & \textbf{1.0059} \\
            \bottomrule
        \end{tabular}
    }
    \label{tab:algos}
\end{table}

Thirteen different cases are tested, none of which are included in the training set. The first row is the ground truth, the second row is the conditional diffusion model (CDM) result and the last row is the result of our method. CASE1-3 contains two inclusions, and both CDM and CDMVC produce satisfactory results. However, as the number of inclusions increases in CASE4-9, accurately reconstructing the angles in triangles and squares becomes more challenging. Specifically, the CDM struggles to reconstructs the exact corner as indicated in red rectangle from CASE5-9. In comparison, CDMVC rectifies incorrect corners more accurately, bringing them closer to the ground truth, although the corners in CASE8 may not be perfectly identical to the ground truth. Furthermore, in the concave model setting, both algorithms achieve relatively good results. Nevertheless, as observed in CASE1-3, CDMVC further improves imaging quality.
\begin{table*}[h]
    \centering
    \caption{Metrics for Generalization Test}
    \begin{threeparttable}
    \small 
    \begin{tabular}{lcccccccccccc}
        \toprule
        \multirow{2}{*}{Algorithm} & \multicolumn{6}{c}{CASE 1} & \multicolumn{6}{c}{CASE 2} \\
        \cmidrule(lr){2-7} \cmidrule(lr){8-13}
         & RE & SSIM & PSNR & {MSE} & {CC} & DR* & RE & SSIM & PSNR & {MSE} & {CC} & DR* \\
        \midrule
       PDIPM &2.6067 & 0.5422 & 12.0826 & {0.0620} & {0.7197} & 0.7774 & 2.056 & 0.4714 &12.1394 & {0.0612} & {0.7681} & 0.8348 \\
       CNN &0.196 & 0.905 & 23.7941 & {0.0050} & {0.9695} & 0.9974 & 0.1839 & 0.889&23.0221 & {0.0060} & {0.9702} &0.9981\\
        CGAN &0.3795 &0.7966 &18.8507 & {0.0152} & {0.8946} & 1.0398 &0.3872 &0.7744 &18.238 & {0.0170} & {0.8882} & 1.0405 \\
       WGAN & 0.2596 &0.8626 & 21.7846 & {0.0084} &{0.9467} & 1.2045 &0.2436&0.8375 &21.3617 & {0.0097} &{0.9511} & 1.1948 \\
        ${\rm{CS}}{{\rm{D}}^*}$ & 0.495 & 0.736 & 20.657 & {0.0205} & {0.8015} & 1.2539 &0.354 &0.753 & 19.3561 & {0.0219} & {0.8321} & 1.1593 \\
        \textbf{CDMVC}  & \textbf{0.1543} &  \textbf{0.9296} & \textbf{25.0365} & {\textbf{0.0039}} & {\textbf{0.9743}} & \textbf{1.0047} & \textbf{0.1351} & \textbf{0.9202} & \textbf{24.482} & {\textbf{0.0044}} & {\textbf{0.9773}} & \textbf{1.0049}\\
        \midrule
        \multirow{2}{*}{Algorithm} & \multicolumn{6}{c}{CASE 3} & \multicolumn{6}{c}{CASE 4} \\
        \cmidrule(lr){2-7} \cmidrule(lr){8-13}
         & RE & SSIM & PSNR & {MSE} & {CC} & DR* & RE & SSIM & PSNR & {MSE} & {CC} & DR* \\
        \midrule
        PDIPM & 1.9297& 0.6528&11.0321 &0.0671 &0.7687 &1.7466 &1.8838 & 0.5146&7.9783 &0.0751 &0.6374 &2.8737 \\
       CNN & 0.5323& 0.8750& 19.1444& 0.0122&0.8436 & 0.9834& 0.2277 & 0.9122& 21.9143& 0.0064&0.9492 &0.9999 \\
        CGAN &0.7347 &0.7274 & 17.1143& 0.8039 &0.0194 &1.1844 &0.3647 & 0.8313& 18.3720& 0.0145 &0.8805& 1.0373\\
       WGAN &0.4765 & 0.8554&19.2679 &0.0118 & 0.8916&1.4201 & 0.2865& 0.8888&20.9312 &0.0081 &0.9359 & 1.1738 \\
        ${\rm{CS}}{{\rm{D}}^*}$ &1.2364  &0.4392 &10.329 &0.0963 &0.5125 &1.641 &1.3547 &0.5283 &9.6193 &0.0894 &0.5781 &1.4382 \\
        \textbf{CDMVC} & \textbf{0.3346} &\textbf{0.9031} & \textbf{ 21.6711}&\textbf{0.0068} & \textbf{0.9165}&\textbf{1.0119} &\textbf{0.1147} &\textbf{0.9724} &\textbf{24.5707} &\textbf{0.0035} & \textbf{0.9808}&\textbf{1.0018} \\
        \midrule
        \multirow{2}{*}{Algorithm} & \multicolumn{6}{c}{CASE 5} & \multicolumn{6}{c}{CASE 6} \\
        \cmidrule(lr){2-7} \cmidrule(lr){8-13}
         & RE & SSIM & PSNR & {MSE} & {CC} & DR* & RE & SSIM & PSNR & {MSE} & {CC} & DR* \\
        \midrule
        PDIPM & 2.4717& 0.5328&11.8825 &{0.0627} & {0.7463}& 0.7572 & 1.3455 & 0.5187&12.0303 &  {0.0627} & {0.7990} & 0.9189\\
       CNN & 0.2092 &0.8784 & 21.3058 & {0.0074} & {0.9504} & 0.9982 & 0.2022 &0.8346&20.3681 & {0.0093} &{0.9502}& 0.9999\\
        CGAN & 0.4111&0.7847& 18.2119 & {0.0180} &{0.8764} &1.0413 & 0.4145 &0.7461&17.5023&{0.0197} & {0.8757} & 1.0454 \\
       WGAN &0.3266 & 0.8411 & 20.1078 & {0.0142} & {0.9115} &1.2712& 0.2867 &0.7605 & 18.0157 & {0.0159} & {0.9290} & 1.2614 \\
        ${\rm{CS}}{{\rm{D}}^*}$ &0.4432& 0.765 & 18.945 & {0.0241} & {0.8202} &1.2368 &0.4877 &0.74 &18.4132 & {0.0207} & {0.8075} & 1.1997 \\
        \textbf{CDMVC}  & \textbf{0.1577} &  \textbf{0.9125} & \textbf{23.6578} & {\textbf{0.0056}} & {\textbf{0.9686}} & \textbf{1.0099} & \textbf{0.1435} &  \textbf{0.8862} & \textbf{ 21.7628} & {\textbf{0.0068}} & {\textbf{0.9698}} & \textbf{1.0053}\\
        
        \bottomrule
    \end{tabular}
    \end{threeparttable}
    \label{tab:generalization data}
\end{table*}

\subsection{Results and Analysis on Test Set With Noise}
\begin{figure}[htpb]
	\centering
	\includegraphics[width=\columnwidth]{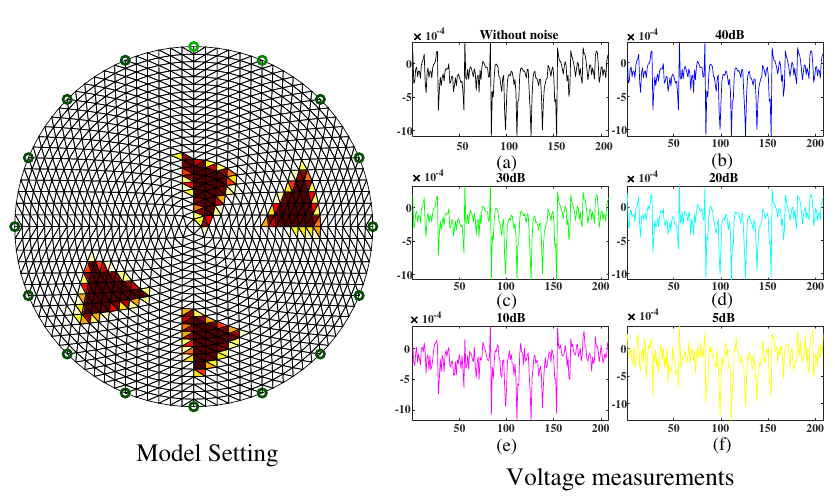}
	\caption{Voltage measurements at different noise level}
	\label{noise_data}
\end{figure}
In order to analyze the anti-noise performance of CDMVC trained with samples without noise, Gaussian noise are added into voltages of the test sets, which is measured by the signal-to-noise ratio (SNR) below:
\begin{equation}
\label{eq:SNR}
SNR =20 \lg \left(\frac{V_{\text {signal }}}{V_{\text {noise }}}\right),
\end{equation}
where $V_{\text {signal }}$ and $V_{\text {noise }}$ represent the effective value of the amplitude of the signal and noise, respectively. A smaller SNR value indicates greater noise intensity. 
Different noise levels ranging from 50 dB to 5 dB are considered. For each noise level, 1300 samples are generated for 13 model settings. Voltage measurements for different noise level are displayed in Fig. \ref{noise_data}. 

\begin{figure}[htpb]
	\centering
	\includegraphics[width=\columnwidth]{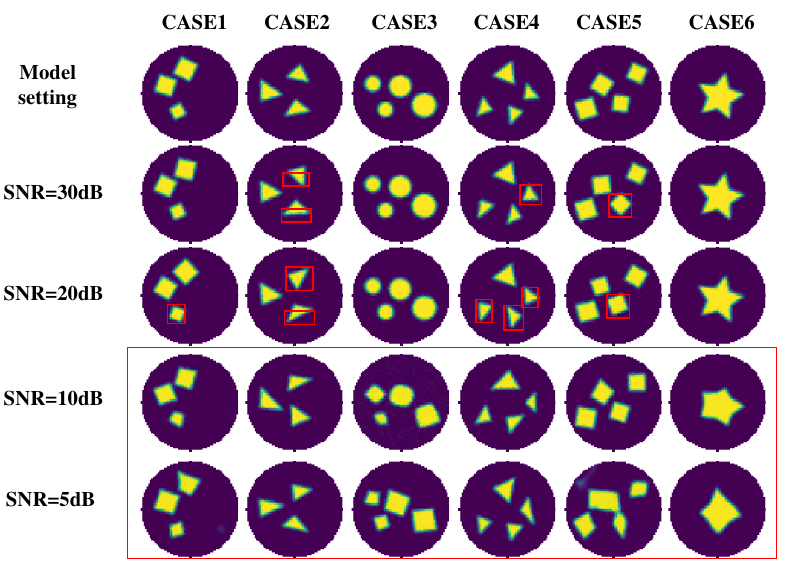}
	\caption{{Visual results under different level noise}}
	\label{anti-noise-vis}
\end{figure}

To evaluate the impact of the number of inclusions on the results, we analyzed metrics corresponding to different quantities of inclusions separately. The statistical metrics for varying inclusion counts are presented in Fig. \ref{anti-noise status}. As the number of inclusions increases, the performance gradually degrades. With increasing noise, the RE metric rises, while both SSIM and PSNR decrease across all model settings. Notably, RE remains below 0.2, and SSIM and PSNR stay above 0.9 and 29, respectively. Additionally, the DR metric remains stable, hovering close to 1.

{The visual results produced by our method are shown in Fig. \ref{anti-noise-vis}. It is worth noting that visual results for noise levels between 50 dB and 40 dB are omitted, as they closely resemble the ground truth. At the 30 dB noise level, slight distortions in the reconstructions are observed, while at 20 dB, the shapes are mildly deformed. At noise levels of 10 dB and 5 dB, the reconstructions show significant shape distortion, highlighted by the red rectangles. Despite these distortions, the locations remain accurate, and most original shapes are still discernible. Both the visual and quantitative metrics demonstrate the robust anti-noise capability of CDMVC.}

\subsection{Evaluation of Model Generalization}
\label{generalization subsection}
To test the generalization of our method, we generated two kinds of samples: 1) samples with new shapes, and 2) samples with complex distributions. Specifically, 100 samples were generated for each situation, resulting in 400 samples in total.

\begin{figure}[htpb]
	\centering
	\includegraphics[width=\columnwidth]{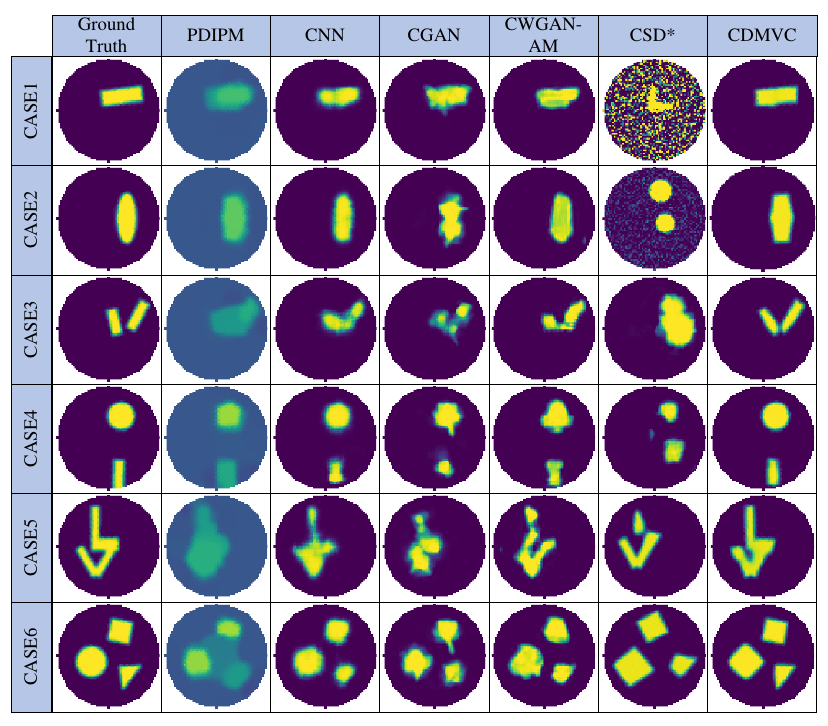}
	\caption{{Visual results of the generalization test.}}
	\label{generalization}
\end{figure}

As shown in Fig. \ref{generalization}, samples with new shapes are represented by CASE1-3. The images constructed by comparative methods exhibit serious artifacts. The CSD$^{*}$ method, in particular, fails to obtain the correct shapes due to the lack of guidance information. In contrast, our method can accurately reconstruct the corners of rectangles and produce clearer inclusion boundaries.
Samples with complex distributions are represented by CASE4-6. In these cases, inclusion boundaries in the reconstructed images are blurred except for those produced by CSD$^{*}$ and our method. However, the shapes reconstructed by CSD$^{*}$ are incorrect, whereas our method successfully captures the overall accurate sizes and shapes. This indicates that our method has the potential to reconstruct more complex distributions.
The detailed metrics for the two types of test samples are shown in TABLE \ref{tab:generalization data}. Our method achieves the lowest RE and MSE, with SSIM values exceeding 0.9, PSNR values above 22, CC values greater than 0.96, and DR values closest to 1, indicating its strong generalization capability.

\section{PHYSICAL EXPERIMENT}
\label{physical experiment}
\begin{figure}[htpb]
	\centering
	\includegraphics[width=\columnwidth]{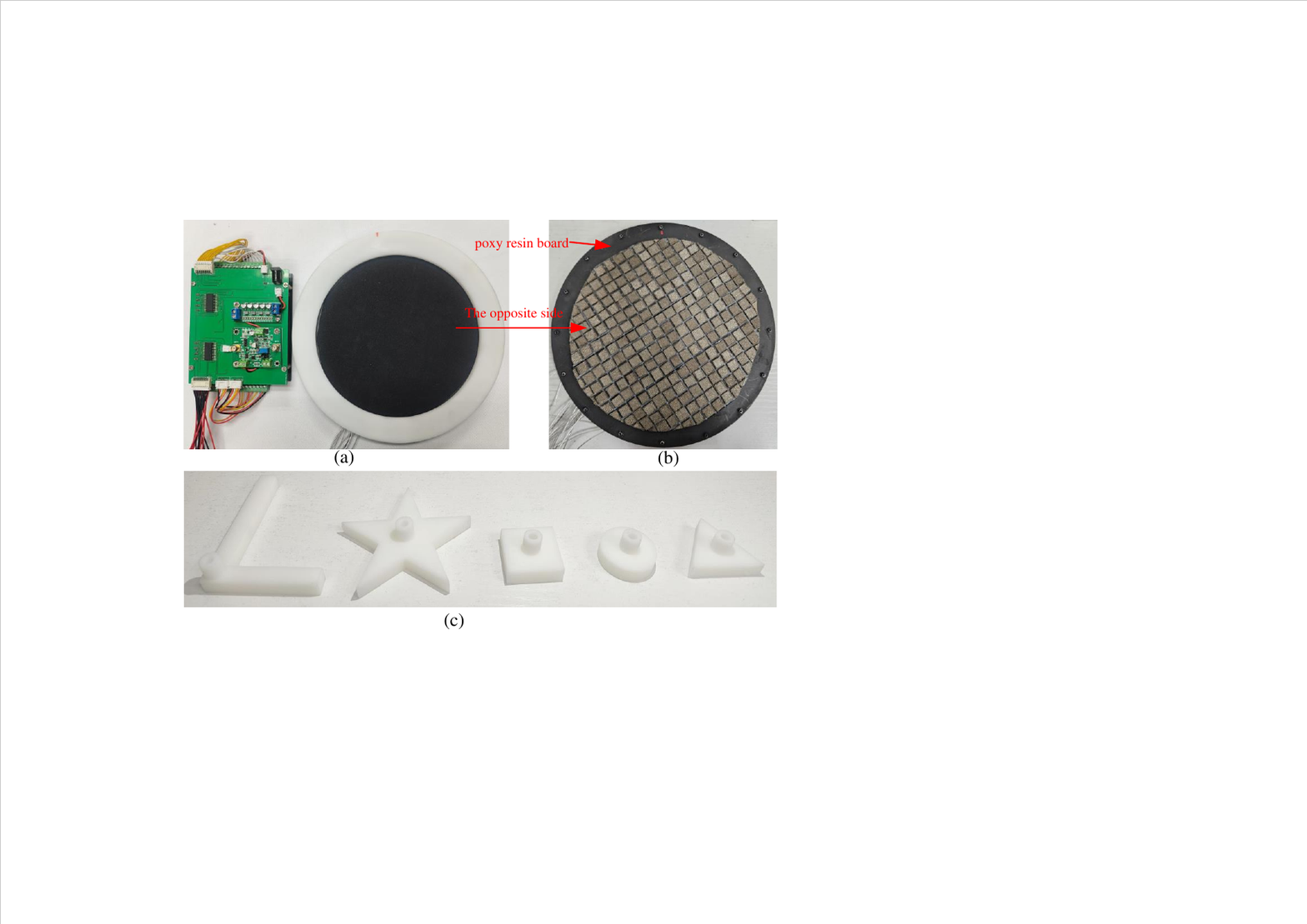}
	\caption{The tactile sensor, data collector and measured objects.}
	\label{device}
\end{figure}


To validate the effectiveness of our method, some physical experiments are conducted with the fabricated tactile sensor. For all experiments, normal force is applied to the sensor.

\subsection{Sensor Fabrication Procedures}
To construct the tactile sensor, a structure consisting of a rigid layer and a flexible layer is adopted. The diameter of rigid layer is 19 cm. It uses an epoxy resin board as the substrate on which the conductive graphite spray (KONTAKT GRAPHIT 33, Germany) is evenly applied as shown in Fig. \ref{device}(a). The flexible layer consists of discrete, non-connected square fabric pieces. Specifically, this layer requires attaching discrete high-conductivity fabric pieces (Silver ﬁber, YSILVER82, China) onto a neoprene foam substrate. Each conductive fabric piece is cut to a size of 8×8 mm², as shown in Fig. \ref{device}(b). Additionally, the measured objects are 3D printed resin which is present in Fig. \ref{device}(c). For more details, please refer to our previous work \cite{zheng2023adaptive}.

\subsection{EIT System Electronics}

\begin{figure}[htpb]
	\centering
	\includegraphics[width=\columnwidth]{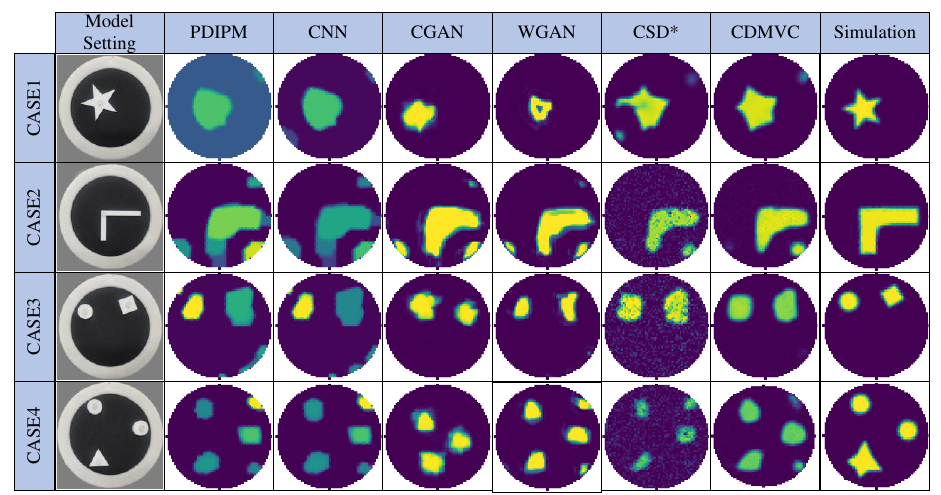}
	\caption{{Visual results of the physical experiment.}}
	\label{experiment}
\end{figure}

To realize tactile sensing on the above fabricated sensor, the EIT experimental system featuring 16 electrodes is illustrated in Fig. \ref{device}. An excitation current with an amplitude of 25 mA is generated by a voltage-controlled current source and injected into the measurement domain through the excitation channel. The analog voltage signal from the measurement channel is digitized by an analog-to-digital converter (ADC). The digital signal is processed by the field-programmable gate array (FPGA) (Cyclone IV EP4CE10F17C8N) and stored in a first in, first out (FIFO) buffer. The STM32F107VCT6 microcontroller handles two main tasks: it controls the multiplexer for channel switching and the ADC chip for analog-to-digital conversion.
\begin{table*}[h]
    \centering
    \caption{{Metrics for Physical Experiment Test}}
    \begin{threeparttable}
    \small 
    \begin{tabular}{lcccccccccccc}
        \toprule
        \multirow{2}{*}{Algorithm} & \multicolumn{6}{c}{CASE 1} & \multicolumn{6}{c}{CASE 2} \\
        \cmidrule(lr){2-7} \cmidrule(lr){8-13}
         & RE & SSIM & PSNR & {MSE} & {CC} & DR* & RE & SSIM & PSNR & {MSE} & {CC} & DR* \\
        \midrule
       PDIPM &2.064 & 0.4425 & 11.561 & {0.0722} & {0.5023} & 0.6453 & 2.190 & 0.3984 &10.8921 & {0.0801} & {0.6499} & 0.9021 \\
       CNN  &1.8082 & 0.4013 & 11.790 & {0.0690} & {0.5344} & 0.7521 & 1.814 & 0.4573 &11.4358 & {0.0790} & {0.6988} & 0.9152 \\
        CGAN &1.947 &0.3866 &10.6571 & {0.0845} & {0.4946} & 1.0257 &0.4983 &0.6057 &15.459 & {0.0481} & {0.7161} & 1.0380 \\
       WGAN & 2.3321 &0.3747 & 9.9701 & {0.0874} &{0.4453} & 1.2336 &0.3978&0.7353 &17.3715 & {0.0391} &{0.8433} & 1.1806 \\
        ${\rm{CS}}{{\rm{D}}^*}$ & 0.954 & 0.8440 & 17.443 & {0.0145} & {0.8835} & 1.2659 &0.4324 &0.673 & 14.7631 & {0.0219} & {0.7420} & 1.2901 \\
        \textbf{CDMVC}  & \textbf{0.7942} &  \textbf{0.8858} & \textbf{19.7741} & {\textbf{0.0092}} & {\textbf{0.9001}} & \textbf{1.0126} & \textbf{0.2457} & \textbf{0.8549} & \textbf{18.549} & {\textbf{0.0184}} & {\textbf{0.8864}} & \textbf{1.0121}\\
        
        \midrule
        \multirow{2}{*}{Algorithm} & \multicolumn{6}{c}{CASE 3} & \multicolumn{6}{c}{CASE 4} \\
        \cmidrule(lr){2-7} \cmidrule(lr){8-13}
         & RE & SSIM & PSNR & {MSE} & {CC} & DR* & RE & SSIM & PSNR & {MSE} & {CC} & DR* \\
        \midrule
        PDIPM & 2.201& 0.3905&10.4779 &{0.0841} & {0.4470}& 0.6398 & 2.0312 & 0.4766&11.9329 &  {0.0653} & {0.5234} & 0.8147\\
       CNN & 1.9450 &0.3844 & 11.0225 & {0.0749} & {0.4584} & 0.6478 & 1.8632 &0.4983&12.3741 & {0.0609} &{0.5344}& 0.8609\\
        CGAN & 1.7657&0.4347& 12.6329 & {0.0673} &{0.5874} &1.0564 & 1.5178 &0.4471&11.7843&{0.0679} & {0.5078} & 1.0790 \\
       WGAN &1.6784 & 0.4129 & 10.8647 & {0.0782} & {0.4658} &1.3348& 0.2978 &0.8319 & 16.5640 & {0.0389} & {0.8890} & 1.0893 \\
        ${\rm{CS}}{{\rm{D}}^*}$ &0.2956& 0.9193 & 15.6014 & {0.0316} & {0.7832} &1.3568 &0.7152 &0.4316 &10.1670 & {0.0896} & {0.4729} & 1.6348 \\
        \textbf{CDMVC}  & \textbf{0.1013} &  \textbf{0.8432} & \textbf{19.8476} & {\textbf{0.0183}} & {\textbf{0.8903} }& \textbf{1.0172} & \textbf{0.1697} &  \textbf{0.8623} & \textbf{ 18.7791} & {\textbf{0.0157}} & {\textbf{0.8915}} & \textbf{1.0116}\\
        \bottomrule
    \end{tabular}
    \end{threeparttable}
    \label{tab:experiment data}
\end{table*}
\begin{table}[htpb]
    \caption{{ Comparison of imaging time for different methods.}}
    \resizebox{\columnwidth}{!}{
        \begin{tabular}{lcccccc}
            \toprule
             \textbf{Method} & \textbf{CNN} & \textbf{CGAN} & \textbf{CWGAN-AM} & \textbf{CSD} & \textbf{CDMVC} \\
            \midrule
           \textbf{Imaging time (s)} & 0.005 & 0.007 & 0.013 & 8.222 & 0.485 \\
            \bottomrule
        \end{tabular}
    }
    \label{tab:imaging_time}
\end{table}

The results from the tactile sensor experiments are shown in Fig. \ref{experiment} and TABLE \ref{tab:experiment data}. {The evaluation demonstrates that our method consistently outperforms other approaches across all key metrics. Specifically, it achieves the lowest RE values in all cases, indicating superior reconstruction accuracy. In terms of SSIM and PSNR, our method yields significantly higher scores compared to other algorithms, highlighting its effectiveness in preserving fine details and minimizing noise. The lower MSE and higher CC values further underscore the method's ability to reduce pixel-wise errors while maintaining a strong correlation with the ground truth images. Additionally, the DR* metric, which evaluates dynamic range preservation, shows that our approach strikes a better balance between noise suppression and detail retention across different scenarios. Together, these results confirm the robustness and generalization capability of our method, especially in handling noisy and complex image structures.}

Similar to the simulation results, reconstructions using PDIPM exhibit significant limitations, such as pronounced artifacts and poor boundary accuracy. The CNN method, being heavily reliant on the initial reconstruction, produces results that closely resemble the initial input, limiting its flexibility. While GAN-based methods offer improved image quality compared to PDIPM and CNN, they still exhibit noticeable discrepancies between the reconstructions and the ground truth.

In contrast, our method demonstrates clear advantages, particularly in addressing complex scenarios. The output images produced by our approach feature sharp boundaries and better preservation of intricate features, especially at corners. This is largely attributed to the voltage consistency constraints integrated within the conditional diffusion model, which effectively preserves both sharp and complex features in the reconstructions. {To be noted, the simulation experiment are conducted corresponding to the physical experiment in the last column of Fig. \ref{experiment}. It shows that there is still a gap between the simulation and actual utilization, but our method can still reconstruct image with high quality.} {Besides, the average imaging time is shown in TABLE \ref{tab:imaging_time}.  Although the imaging time of the proposed CDMVC method is longer than that of the CNN, CGAN, and CWGAN-AM methods, it is markedly faster than the CSD* method. In contrast, our method significantly reduces computational costs while maintaining high reconstruction accuracy. This demonstrates that our approach strikes an optimal balance between computational efficiency and performance. As a result, it serves as a practical solution for real-time applications where both speed and high-quality reconstructions are essential. This is particularly important in scenarios characterized by noisy or complex data.}

\section{Conclusion}
\label{conclusion}

A conditional diffusion model with voltage consistency is proposed for EIT problem in this study. CDMVC consists of a pre-imaging module, a conditional diffusion model for reconstruction, a forward voltage constraint network and a scheme of voltage consistency constraint during sampling process. The pre-imaging module provide the prior about the shape and position of inclusions. The initial reconstruction-based diffusion model takes the prior and further obtains the knowledge of different shapes. To incorporate physical information, the forward voltage constraint network is designed which conduct the voltage consistency constraint during the sampling process. To demonstrate the effectiveness of our method, comprehensive experiments are conducted. The reconstruction results show that our has good anti-noise ability. Compared with traditional method, CNN method, GAN-based methods and method based on diffusion models, our method can reconstruct the irregular boundaries and shape/size variations of complex inclusions more accurately.

{Our method focuses on post-processing the initial reconstruction to enhance image quality. Compared to other similar post-processing approaches, it exhibits a lower dependency on the initial reconstruction. Besides, it is capable of generating robust reconstructions even when the initial data is distorted. Our approach also entails a level of computational complexity, but the requirement of real-time imaging can be satisfied.
Future work will explore optimizing the computational efficiency of the model to enable its application in real-time scenarios. Another avenue for future research includes scaling the method to large datasets, which may require exploring new technology like rectified flow that focus on bridging arbitrary distribution.}

\section*{Acknowledgments}
This work were supported by the National Natural Sci- ence Fund for Key International Collaboration under Grant 62120106005 and the National Natural Science Foundation of China under Grant 62273054, 62303259.

\bibliographystyle{IEEEtran}
\bibliography{reference}

\end{document}